\documentclass{article} 
\usepackage{iclr2018_conference,times}
\usepackage{hyperref}
\usepackage{url}

\usepackage{algorithm}
\usepackage{algorithmic}
\usepackage{amsmath}
\usepackage{amsfonts}
\usepackage{graphics}
\usepackage{epsfig}
\usepackage{subfigure}
\usepackage{float}
\usepackage{multirow}
\usepackage{appendix}
\usepackage{color}
\usepackage{times}
\usepackage{helvet}
\usepackage{courier}
\usepackage{multirow}
\usepackage{bm}
\usepackage[american]{babel}

\newtheorem{prop}{Proposition}[section]
\newtheorem{cor}{Corollary}[section]

\def \perm{\text{perm}}
\def \L{\mathcal{L}}

\def \u{{\bf u}}
\def \v{{\bf v}}
\def \w{{\bf w}}
\def \x{{\bf x}}
\def \y{{\bf y}}
\def \z{{\bf z}}
\def \b{{\bf b}}
\def \d{{\bf d}}
\def \c{{\bf c}}
\def \a{{\bf a}}

\def \hw{\hat{\bf{w}}}
\def \W{{\bf W}}
\def \D{{\bf D}}
\def \H{{\bf H}}
\def \X{{\bf X}}
\def \I{{\bf I}}

\def \m{{\bf m}}
\def \v{{\bf v}}
\def \s{{\bf s}}
\def \p{{\bf p}}
\def \q{{\bf q}}
\def \S{{\mathcal S}}
\def\CE{\mathbf{E}}

\def\diag{\text{diag}}

\newcommand{\NM}[2 ]{\left\| #1 \right\|_{#2}}

\iclrfinalcopy

\title{Loss-aware Weight Quantization of Deep Networks}

\author{Lu Hou, 
	James T. Kwok 
	\\
	Department of Computer Science and Engineering\\
	Hong Kong University of Science and Technology\\
	Hong Kong \\
	\texttt{\{lhouab,  jamesk\}@cse.ust.hk} \\
}

\begin{document}

\maketitle

\begin{abstract}
The huge size of
deep networks 
hinders their use in small computing devices.
In this paper, we consider compressing the network by weight quantization. We extend a
recently proposed loss-aware weight binarization scheme to ternarization, with possibly
different scaling parameters for the positive and negative weights, and
$m$-bit 
(where $m>2$) quantization.
Experiments on feedforward and recurrent neural networks show that the proposed 
scheme 
outperforms state-of-the-art weight quantization algorithms,
and
is as accurate (or even more accurate) than the full-precision network.
\end{abstract}


\section{Introduction}

The last decade has witnessed huge success of deep neural networks in
various domains. Examples include computer vision, speech recognition, and natural language processing~\citep{lecun2015deep}. 
However, their
huge size
often hinders deployment to small computing devices such as cell phones and the internet of
things.
Many 
attempts have been 
recently
made to reduce the model size.
One common approach is to prune
a trained dense network 
\citep{han2015learning, han2015deep}.  However, 
most
of the pruned weights may 
come from the
fully-connected layers where computations are cheap,
and the resultant time reduction is insignificant.
\citet{li2017pruning} and \citet{molchanov2017pruning}
proposed to prune filters in the convolutional neural networks 
based on their
magnitudes or
significance to the loss. However, the pruned network has to be retrained, which is again expensive.
	
Another direction is to use 
more compact
models.  GoogleNet~\citep{szegedy2015going} and ResNet~\citep{he2016deep} replace the
fully-connected layers with simpler global average pooling. However, they are also deeper.
SqueezeNet~\citep{iandola2016squeezenet} reduces the model size by replacing most of the
$3\times3$ filters with $1 \times 1$ filters.
This is less efficient on smaller networks because the 
dense $1\times1$ convolutions
are costly.
MobileNet~\citep{howard2017mobilenets} compresses the model using separable depth-wise
convolution.
ShuffleNet~\citep{zhang2017shufflenet}  utilizes pointwise group convolution and channel
shuffle to reduce the computation cost while maintaining accuracy.
However, highly optimized  group convolution and depth-wise convolution implementations are required.
Alternatively, \citet{novikov2015tensorizing}  
compressed the model 
by using a compact multilinear format to represent the dense weight matrix.
The CP 
and Tucker
decompositions
have also been used on the kernel tensor in CNNs 
\citep{lebedev2014speeding,kim2015compression}.
However, they often need expensive fine-tuning.
	
Another effective approach to compress the network and accelerate training
is by quantizing each full-precision weight to a small number of bits.
This can be further divided to two sub-categories, depending on whether pre-trained models are used
\citep{lin2016fixed,mellempudi2017ternary} or the quantized model is trained
from scratch~\citep{courbariaux2015binaryconnect,li2017training}. 
Some of these also directly learn with low-precision weights,
but they usually suffer from severe accuracy deterioration \citep{li2017training,miyashita2016convolutional}.
By keeping the full-precision weights during learning, 
\citet{courbariaux2015binaryconnect} pioneered 
the BinaryConnect algorithm, which uses only one bit for each weight while still achieving state-of-the-art 
classification 
results.
\citet{rastegari2016xnor} further incorporated weight scaling, and obtained better results.
Instead of simply finding the closest binary approximation of the full-precision weights, a loss-aware scheme is proposed in~\citep{hou2017loss}.
Beyond binarization, TernaryConnect \citep{lin2015neural} quantizes each weight to
$\{-1,0,1\}$.  \citet{li2016ternary} and \citet{zhu2017trained} added scaling to the ternarized weights,
and DoReFa-Net~\citep{zhou2016dorefa} further extended quantization to more than three
levels. However, these methods do not 
consider the effect of quantization on the loss, and rely on heuristics in their procedures
\citep{zhou2016dorefa,zhu2017trained}.
Recently, a loss-aware low-bit quantized neural network  is proposed 
in \citep{leng2017extremely}. However,
it uses full-precision weights in the forward pass and 
the 
extra-gradient
method
\citep{vasilyev2010extragradient}
for update,
both of which are expensive.


In this paper, we propose an efficient and disciplined ternarization scheme for network compression.  Inspired by \citep{hou2017loss}, we explicitly consider the effect of ternarization on the loss.
This is formulated as an optimization problem which is then solved efficiently by the proximal Newton algorithm.
When the 
loss surface's
curvature is ignored,
the proposed method reduces to that of \citep{li2016ternary}, and is also related to 
the projection step of~\citep{leng2017extremely}.
Next, we extend it to 
(i) allow the use of different scaling parameters for the positive and negative weights;
and (ii) the use of $m$  bits (where $m>2$) for weight quantization.
Experiments on both feedforward and recurrent neural
networks show that the proposed quantization scheme outperforms state-of-the-art algorithms.



{\bf Notations:}
For a vector $\x$, $\sqrt{\x}$ denotes the element-wise square root (i.e., $[\sqrt{\x}]_i =
\sqrt{x_i}$), $|\x|$ is the element-wise absolute value, $\|\x\|_p =
(\sum_i|x_i|^p)^{\frac{1}{p}}$ is its $p$-norm,
and
$\text{Diag}(\x)$ returns a  diagonal matrix with
$\x$
on the diagonal.
For two vectors $\x$ and $\y$,
$\x \odot \y$ denotes the element-wise multiplication
and $\x \oslash \y$ the element-wise division.
$\|\x\|_{\bm Q}^2 = \x^\top {\bm Q} \x$.
Given a threshold $\Delta$,
$\I_{\Delta}(\x)$ returns a vector 
such that
$[\I_{\Delta}(\x)]_i=1$  if $x_i>\Delta$,  $-1$ if $x_i < -\Delta$, and 0 otherwise.
$\I_{\Delta}^+(\x)$ considers
only the positive threshold,
i.e., $[\I^+_{\Delta}(\x)]_i=1$ if $x_i>\Delta$, and 0 otherwise.
Similarly,
$[\I^-_{\Delta}(\x)]_i=-1$ if $x_i< -\Delta$, and 0 otherwise.
For a matrix $\X$,
$\text{vec}(\X)$ returns a vector by stacking  all
the columns 
of $\X$,
and $\text{diag}(\X)$ returns a vector whose entries are from the diagonal of $\X$.


\section{Related Work}

Let the full-precision weights from all $L$ layers be $\w =
	[\w_1^\top,  \w_2^\top, \dots, \w_L^\top]^\top$, where
	$\w_l =  \text{vec}(\W_l)$, and $\W_l$ is the weight matrix at layer $l$. 
The corresponding quantized 
weights 
will be denoted $\hw=[\hw_1^\top, \hw_2^\top, \dots, \hw_L^\top]^\top$.



\subsection{Weight Binarized Networks}

In
BinaryConnect
\citep{courbariaux2015binaryconnect},
each element of $\w_l$ 
is binarized
to $-1$ or $+1$ by using the sign function:
$\text{Binarize}(\w_l) = \text{sign} (\w_l) $.
In 
the Binary-Weight-Network (BWN) \citep{rastegari2016xnor},
a scaling parameter 
is also included,
i.e., $\text{Binarize}(\w_l) = \alpha_l \b_l$, where 
$\alpha_l>0$,
$\b_l \in \{-1, +1\}^{n_l}$ and
$n_l$ is the number of weights in $\w_l$.
By minimizing the difference between $\w_l$ and $\alpha_l \b_l$,
the optimal 
$\alpha_l, \b_l$ 
have the simple form:
$\alpha_l = \|\w_l\|_1/n_l$, and $\b_l = \text{sign} (\w_l)$.

Instead of simply finding the best binary approximation for the full-precision weight $\w_l^t$
at iteration $t$,
the  loss-aware binarized network (LAB)
directly minimizes the loss w.r.t. the binarized weight
$\alpha^t_l \b_l^t$
\citep{hou2017loss}. 
Let $\d^{t-1}_l$ be a vector containing the diagonal of an approximate Hessian of the loss.
It can be shown that 
$\alpha^t_l = \|\d^{t-1}_l \odot \w^t_l\|_1/\|\d^{t-1}_l\|_1$  and
$\b_l^t = \text{sign}(\w^t_l)$.


\subsection{Weight Ternarized Networks}

In a weight ternarized network, zero is used as an additional quantized value.
In TernaryConnect~\citep{lin2015neural}, 
each weight value is clipped to $[-1,1]$ before quantization, and then
a non-negative
weight  $[\w_l^t]_i$ is stochastically quantized to 
$1$ with probability $[\w_l^t]_i$ (and $0$ otherwise).
When $[\w_l^t]_i$ is negative, it is quantized to 
$-1$ with probability $-[\w_l^t]_i$, and $0$ otherwise.

In 
the ternary weight network (TWN)~\citep{li2016ternary},
$\w_l^t$ is quantized to 
$\hw_l^t = \alpha_l^t \I_{\Delta_l^t}(\w_l^t) $,
where $\Delta_l^t$ is a threshold 
(i.e.,
$[\hw_l^t]_i = 
\alpha_l^t$ if $[\w_l^t]_i>\Delta_l^t$,
$-\alpha_l^t$ if $[\w_l^t]_i<-\Delta_l^t$ and 
0 otherwise).
To obtain 
$\Delta_l^t$ and $\alpha_l^t$,
TWN minimizes the $\ell_2$-distance between the
full-precision and ternarized weights, leading to
\begin{equation} \label{eq:twn}
\Delta_l^t = \arg \max_{\Delta>0} \frac{1}{\|\I_{\Delta}(\w_l^t)\|_1} \left(\sum_{i :
|[\w_l^t]_i|>
\Delta_l^t}
|[\w_l^t]_i| \right)^2, \;\;
\alpha_l^t = \frac{1}{\|\I_{\Delta_l^t}(\w_l^t)\|_1} \sum_{i:
|[\w_l^t]_i|>
\Delta_l^t}
|[\w_l^t]_i|.
\end{equation}
However,
$\Delta_l^t$ 
in (\ref{eq:twn})
is difficult to solve. Instead, TWN simply sets
$\Delta_l^t= 0.7 \cdot \CE(|\w_l^t|)$  in practice.

In TWN, one scaling parameter
($\alpha_l^t$)
is used for both the positive and negative weights at layer $l$.  
In the trained ternary quantization (TTQ) network~\citep{zhu2017trained}, 
different scaling parameters ($\alpha_l^t$  and $\beta_l^t$) are used.
The weight 
$\w_l^t$
is thus quantized to
$\hw_l^t= \alpha_l^t \I_{\Delta_l^t}^+(\w_l^t) + \beta_l^t \I_{\Delta_l^t}^-(\w_l^t)$.
The scaling parameters are learned by gradient descent.
As for $ \Delta_l^t$, two heuristics are used. The first sets $ \Delta_l^t$ to a constant fraction of  $\max(|\w_l^t|) $,  while the second sets $ \Delta_l^t$ such that 
	at all layers
	are equally
sparse.


\subsection{Weight Quantized Networks}

In a weight quantized network, 
$m$ 
bits 
(where $m\ge 2$) 
are used  to represent each weight. 
Let 
$\mathcal{Q}$ be a set of $(2k+1)$ quantized values, where $k=2^{m-1}-1$.
The two popular choices 
of $\mathcal{Q}$ 
are 
$\left\{-1, -\frac{k-1}{k}, \dots, -\frac{1}{k}, 0, \frac{1}{k}, \dots, \frac{k-1}{k}, 1\right\}$
(linear quantization), 
and
$\left\{-1, -\frac{1}{2}, \dots, -\frac{1}{2^{k-1}}, 0, \frac{1}{2^{k-1}}, \dots, \frac{1}{2}, 1\right\}$
(logarithmic quantization).
By limiting the quantized values to powers of two, 
logarithmic quantization is advantageous in that
expensive floating-point operations can be replaced by cheaper bit-shift operations.  
When $m=2$, 
both schemes reduce to $\mathcal{Q}=\{-1, 0, 1\}$.

In the DoReFa-Net~\citep{zhou2016dorefa}, weight
$\w_l^t$
is heuristically quantized to $m$-bit, with:\footnote{Note that the quantized value of 0 is not used in DoReFa-Net.}
\[ [\hw_l^t]_i = 2 \cdot \text{quantize}_m \left( \frac{\tanh([\w_l^t]_i)}{2\max(|\tanh([\w_l^t]_i)|)} + \frac{1}{2} \right)-1 \] 
in $\{-1, -\frac{2^m-2}{2^m-1}, \dots, -\frac{1}{2^m-1}, \frac{1}{2^m-1}, \dots,
\frac{2^m-2}{2^m-1}, 1 \}$,
where
$\text{quantize}_m(x) = \frac{1}{2^m-1}\text{round}((2^m-1)x)$.
Similar to 
loss-aware binarization \citep{hou2017loss}, 
\citet{leng2017extremely}
proposed a loss-aware quantized network  
called low-bit neural network (LBNN).
The alternating direction method of multipliers (ADMM) \citep{boyd-11}
is used 
for optimization.
At the $t$th iteration, 
the full-precision weight $\w_l^t$ 
is first updated
by the method of extra-gradient~\citep{vasilyev2010extragradient}:
\begin{equation} \label{eq:extra}
\tilde{\w}_l^t = \w_l^{t-1} - \eta^t \nabla_l \L(\w_l^{t-1}), 
\;\;
\w_l^t = \w_l^{t-1} - \eta^t \nabla_l \L(\tilde{\w}_l^t),
\end{equation} 
where $\L$ is the augmented Lagrangian in the ADMM formulation, and
$\eta^t$ is the stepsize. Next,
$\w_l^t$ is
projected to the space of
$m$-bit quantized weights
so that $\hw_l^t$ is of
the form $\alpha_l \b_l$,
where $\alpha_l > 0$, and $\b_l\in \left\{-1, -\frac{1}{2}, \dots, -\frac{1}{2^{k-1}}, 0,
\frac{1}{2^{k-1}}, \dots, \frac{1}{2}, 1\right\}$. 


\section{Loss-Aware Quantization}


\subsection{Ternarization using Proximal Newton Algorithm}

In weight
ternarization,
TWN
simply finds the closest ternary approximation of the full precision weight
at each iteration,
while TTQ
sets the ternarization threshold
heuristically. 
Inspired by LAB  (for binarization),
we consider the loss explicitly during quantization and obtain 
the quantization thresholds and scaling parameter by solving an optimization problem.

As in TWN, the weight $\w_l$  
is ternarized as
$\hw_l =\alpha_l \b_l$, where $\alpha_l>0$  and $\b_l\in \{-1, 0, 1\}^{n_l}$.
Given a loss function $\ell$,
we formulate
weight ternarization 
as the following optimization problem:
\begin{equation} \label{eq:obj}
\min_{\hw} \; \ell(\hw) \;:\;
\hw_l=\alpha_l \b_l, \;
\alpha_l > 0,
\;
\b_l\in \mathcal{Q}^{n_l}, \;\;  l=1,\dots,L, 
\end{equation}
where $\mathcal{Q}$ is the set of desired quantized values.
As in LAB, we will 
solve this
using the proximal Newton method~\citep{lee2014proximal,rakotomamonjy2016dc}.
At iteration $t$,
the objective is replaced by the second-order expansion
\begin{equation} \label{eq:2nd}
\ell(\hw^{t-1}) + \nabla \ell(\hw^{t-1})^\top (\hw -
	\hw^{t-1}) + \frac{1}{2}(\hw - \hw^{t-1})^\top \H^{t-1} (\hw - \hw^{t-1}), 
\end{equation} 
where $\H^{t-1}$ is an estimate of the Hessian of $\ell$ at $\hw^{t-1}$.
We use the diagonal equilibration pre-conditioner~\citep{dauphin2015equilibrated}, which is 
robust in the presence of saddle points and also
readily available in popular
stochastic deep network optimizers
such as 
Adam \citep{kingma2014adam}.
Let $\D_l$  be the approximate diagonal Hessian at layer $l$.
We use 
$\D=\text{Diag}([\diag(\D_1)^\top, \dots, \diag(\D_L)^\top]^\top)$
as an estimate of $\H$.
Substituting (\ref{eq:2nd}) into (\ref{eq:obj}), 
we solve the following subproblem at the $t$th iteration:
\begin{eqnarray} \label{eq:obj_proximal}
 &\min_{\hw^t} &   \nabla \ell(\hw^{t-1})^\top (\hw^t - \hw^{t-1})+\frac{1}{2}(\hw^t - \hw^{t-1})^\top \D^{t-1} (\hw^t - \hw^{t-1})  \\ \nonumber
& \text{s.t.} &  \hw_l^t=\alpha_l^t \b_l^t, 
\; \alpha_l^t > 0, \; \b_l^t\in \mathcal{Q}^{n_l},  \;\; l=1,\dots,L. \label{eq:quan}
\nonumber
\end{eqnarray}

	\begin{prop} \label{prop:two_step}
		Let $\d^{t-1}_l \equiv \text{diag}(\D^{t-1}_l)$, the objective in (\ref{eq:obj_proximal}) 
		can be rewritten as
		\begin{equation} \label{eq:ter_quantize}
		\min_{\hw^t}
		\frac{1}{2} \sum_{l=1}^{L} \|\hw_l^t- \w_l^t\|_{\D_l^{t-1}}^2,
		\end{equation}
		where 
		\begin{equation} \label{eq:ter_sgd}
		\w^t_l \equiv \hw_l^{t-1} - \nabla_l \ell(\hw^{t-1}) \oslash \d^{t-1}_l. 
		\end{equation}
	\end{prop}

Obviously, this objective can be minimized layer by layer.
Each proximal Newton iteration 
thus
consists of two steps: 
(i) 
Obtain $\w^t_l$ in (\ref{eq:ter_sgd}) 
by gradient descent along $ \nabla_l \ell(\hw^{t-1})$, which is
preconditioned by the adaptive learning rate $1 \oslash \d^{t-1}_l$ so that the rescaled dimensions have similar curvatures;
(ii) Quantize $\w^t_l$ to $\hat{\w}^t_l$ by
minimizing the scaled difference 
between
$\hat{\w}_l^t$ and $ \w_l^t$
	in (\ref{eq:ter_quantize}).
Intuitively, 
when the curvature is low ($[\d_l^{t-1}]_i$ is small),  the loss is not sensitive
to the weight  and ternarization error can be less penalized.
When 
the loss surface is steep,
ternarization 
has to be more accurate.

Though the constraint in 
(\ref{eq:quan})
is more complicated than that in LAB, interestingly the following simple relationship can still be obtained
for weight ternarization. 

\begin{prop} \label{prop:ter_alt}
With 
$\mathcal{Q}=\{-1, 0, 1\}$,
and the optimal 
$\hat{\w}_l^t$ 
in (\ref{eq:ter_quantize}) 
of the form $\alpha \b$.
For a fixed $\b$,
$\alpha = \frac{\|\b \odot \d_l^{t-1} \odot \w_l^t\|_1}{\|\b \odot  \d_l^{t-1}\|_1}$;
whereas when
$\alpha$ is
fixed,
$\b = \I_{\alpha/2}(\w_l^t)$.
\end{prop}
Equivalently,
$\b$ can be written as  
$\Pi_{\mathcal{Q}} (\w_l^t/\alpha)$,
where $\Pi_{\mathcal{Q}} (\cdot)$ projects each entry of the input argument to the nearest
element in $\mathcal{Q}$.
Further discussions on
how to solve for $\alpha_l^t$
will  be presented in 
Sections~\ref{sec:exact} and \ref{sec:approx}.
When the curvature is the same for all dimensions at layer $l$, 
the following Corollary shows that 
the solution above reduces that   
of TWN.
\begin{cor} 
	\label{cor:twn}
When $\D_l^{t-1} = \lambda \I$,
$\alpha_l^t$ 
reduces to 
the TWN solution in (\ref{eq:twn}) 
with 
$\Delta_l^t = \alpha_l^t/2$.
\end{cor} 

In other words, TWN
corresponds to using the proximal gradient algorithm, while the proposed method corresponds
to using the proximal Newton algorithm with diagonal Hessian.  In composite optimization, it
is known that the proximal Newton algorithm is more efficient than the
	proximal gradient algorithm~\citep{lee2014proximal,rakotomamonjy2016dc}. 
Moreover, note that the interesting relationship 
$\Delta_l^t = \alpha_l^t/2$
is not observed in TWN, while
TTQ  
completely neglects this relationship.

In LBNN~\citep{leng2017extremely},
its projection step 
uses an objective 
which
is similar to (\ref{eq:ter_quantize}),
but without using
the curvature information. Besides,
their
$\w_l^t$ is updated with the extra-gradient in (\ref{eq:extra}),
which doubles the number of forward, backward and update steps, and can be costly. 
Moreover, 
LBNN uses full-precision weights
in the forward pass, while
all other quantization methods including ours use quantized
weights (which eliminates most of the multiplications and thus faster training).  

When (i) 
$\ell$ is continuously differentiable with Lipschitz-continuous
gradient (i.e., there exists $\beta>0$ such that $\NM{\nabla \ell(\u) - \nabla \ell(\v)}{2}
\leq \beta \NM{\u-\v}{2}$ for any $\u, \v$); (ii) $\ell$ is bounded from below; and (iii)
$[\d_l^t]_k > \beta \; \forall l,k,t$,
it can be shown that the objective of (\ref{eq:obj}) produced by
the proximal Newton algorithm (with 
solution in Proposition~\ref{prop:ter_alt}) converges
\citep{hou2017loss}.
In practice, 
it is important to keep the full-precision weights during update
\citep{courbariaux2015binaryconnect}.
Hence, we replace (\ref{eq:ter_sgd}) by
$\w^t_l  \leftarrow \w^{t-1}_l  -  \nabla_l \ell(\hw^{t-1}) \oslash \d_l^{t-1} $.
The whole procedure, which is called Loss-Aware Ternarization (LAT), is shown in Algorithm~\ref{alg:whole} of
Appendix~\ref{apdx:whole}. It is similar to Algorithm~1 of LAB \citep{hou2017loss},
except that 
$\alpha^t_l$ and $\b_l^t$ are 
computed
differently.
In step 4, following~\citep{li2016ternary},
we first rescale input 
$\x_l^{t-1}$ 
with $\alpha_l$, so that multiplications in dot products
and convolutions become additions. 
Algorithm~\ref{alg:whole} can also be easily extended
to ternarize 
weights in recurrent networks.
Interested readers are referred to
\citep{hou2017loss} for details.


\subsubsection{Exact solution of $\alpha_l^t$}
\label{sec:exact}

To simplify notations, we drop the superscripts and subscripts. 
From Proposition~\ref{prop:ter_alt},
\begin{equation} \label{eq:opt_exact}
\alpha =  \frac{\|\b\odot \d \odot \w\|_1}{\|\b\odot \d\|_1},
\;\;
\b = \I_{\alpha/2}(\w).
\end{equation}
We now consider how to solve for $\alpha$. 
First, we introduce some notations. 
Given a vector $\x 
=[x_1, x_2, \dots, x_n]$,
and an indexing vector $\s \in \mathbb{R}^n$ whose
entries are a permutation of $\{1, \dots, n\}$, $\perm_\s(\x)$
returns the vector 
$[x_{s_1}, x_{s_2}, \dots x_{s_n}]$,
and $\text{cum}(\x)= 
[x_1, \sum_{i=1}^{2}x_i, \dots, \sum_{i=1}^n x_i]$
returns partial sums for elements in $\x$.
For example, let $\a = [1, -1, -2]$, and $\b=[3, 1, 2]$. Then, $\perm_\b(\a)
= [-2, 1,-1]$
and $\text{cum}(\a) = [1, 0, -2]$.

We sort elements of $|\w|$ in descending order, and let the vector containing the sorted
indices be
$\s$.
For example,
if $\w=[1, 0, -2]$, then
$\s = [3, 1, 2]$.
From (\ref{eq:opt_exact}),
\begin{equation} \label{eq:tmp1}
\alpha 
= \frac{\|\I_{\alpha/2}(\w) \odot \d \odot \w\|_1}{\|\I_{\alpha/2}(\w) \odot \d\|_1}
=  \frac{[ \text{cum}(\perm_{\s}( |\d \odot \w|))]_j}{[
\text{cum}(
\perm_{\s}(
|\d|))]_j} = 2c_j,
\end{equation} 
where $\c = \text{cum}(\perm_{\s}(|\d \odot \w|)) \oslash \text{cum}(\perm_{\s}(\d)) \oslash
2$,
and $j$ is the index  such that
\begin{equation} \label{eq:j}
[\perm_{\s}(|\w|)]_j > c_j> [\perm_{\s}(|\w|)]_{j+1}.
\end{equation} 
For simplicity of notations, let the dimensionality of $\w$ (and thus also of $\c$) be 
$n$,
and
the operation 
$\text{find}(\text{condition}(\x))$ 
returns all indices in $\x$ that satisfies the condition.
It is easy to see that any $j$ satisfying
(\ref{eq:j}) is in
$\S\equiv \text{find}([\perm_{\s}(|\w|)]_{[1:(n-1)]}-\c_{[1:(n-1)]}) \odot ([\perm_{\s}(|\w|)]_{[2:n]} -
	\c_{[1:n-1]})< 0 )$, where
$\c_{[1:(n-1)]}$ is the subvector of $\c$ with elements in the index range 1 to $n-1$.
The optimal $\alpha$ ($=2c_j$) is then the one which yields the smallest objective
in (\ref{eq:ter_quantize}), which can be simplified by
Proposition	~\ref{prop:global_alpha} below.
The procedure is shown in Algorithm~\ref{alg:exact}. 

\begin{prop} \label{prop:global_alpha}
The optimal $\alpha_l^t$ of (\ref{eq:ter_quantize}) equals
$2\arg \max_{c_j: j \in \S} c_j^2 \cdot [\text{cum}(\perm_{\s}(\d_l^{t-1}))]_j$.
\end{prop}

\begin{algorithm}
\caption{Exact solver of (\ref{eq:ter_quantize})}\label{alg:exact}
\begin{algorithmic}[1]
\STATE{\bf Input:} full-precision weight $\w_l^t$, diagonal entries of the approximate Hessian
$\d_l^{t-1}$.
\STATE $\s = \arg \text{sort} (|\w_l^t|)$;
\STATE $\c = \text{cum}(\perm_{\s}(|\d_l^{t-1} \odot \w_l^t|)) \oslash \text{cum}(\perm_{\s}(\d_l^{t-1})) \oslash 2$;
\STATE $\S = \text{find}(([\perm_{\s}(|\w_l^t|)]_{[1:(n-1)]}-\c_{[1:(n-1)]}) \odot ([\perm_{\s}(|\w_l^t|)]_{[2:n]} - \c_{[1:n-1]}) < 0)$;
\STATE $\alpha_l^t = 2\arg \max_{c_j: j \in \S} c_j^2 \cdot [\text{cum}(\perm_{\s}(\d_l^{t-1}))]_j$;
\STATE $\b_l^t = \I_{\alpha_l^t/2}(\w_l^t)$;
\STATE{\bf Output:} $\hw_l^t = \alpha_l^t \b_l^t$.
	\end{algorithmic}
\end{algorithm}


\subsubsection{Approximate solution of $\alpha_l^t$}
\label{sec:approx}

In case the sorting operation 
in step~2 
is expensive,
$\alpha_l^t$ and $\b_l^t$ 
can 
be obtained by alternating the iteration
in Proposition~\ref{prop:ter_alt}
(Algorithm~\ref{alg:approx}).
Empirically, it
converges very fast, usually 
in 5 iterations.

\begin{algorithm}
\caption{Approximate solver for (\ref{eq:ter_quantize}).}
\label{alg:approx}
\begin{algorithmic}[1]
\STATE{\bf Input:} $\b_l^{t-1}$, full-precision weight $\w_l^t$,   diagonal entries of the approximate Hessian $\d_l^{t-1}$.
\STATE{\bf Initialize:} $\alpha = 1.0, \alpha_{\text{old}} = 0.0, \b=\b_l^{t-1}$, $\epsilon=10^{-6}$;
		\WHILE{$|\alpha - \alpha_{\text{old}}|>\epsilon$}
		\STATE $\alpha_{\text{old}} = \alpha$;
		\STATE 
		$\alpha = \frac{\|\b \odot \d_l^{t-1} \odot \w_l^t\|_1}{\|\b \odot \d_l^{t-1}\|_1}$;
		\STATE $\b = \I_{\alpha/2}(\w_l^t)$;
		\ENDWHILE
		\STATE{\bf Output:} 
		$\hw_l^t = \alpha \b$.
	\end{algorithmic}
\end{algorithm}


\subsection{Extension to Ternarization with Two Scaling Parameters}

As in TTQ \citep{zhu2017trained}, we can use different scaling parameters 
for the positive and negative weights in each layer.
The  optimization subproblem at the $t$th iteration then becomes:
\begin{eqnarray} \label{eq:obj_two}
&\min_{\hw^t} &   \nabla \ell(\hw^{t-1})^\top (\hw^t - \hw^{t-1}) + \frac{1}{2}(\hw^t -\hw^{t-1})^\top \D^{t-1} (\hw^t - \hw^{t-1})  \\ \nonumber
& \text{s.t.} &  \hw_l^t \in \{-\beta_l^t, 0, \alpha_l^t\}^{n_l}, \; \; \alpha_l^t>0,  \;\; \beta_l^t>0, \; l = 1, \dots, L.
\nonumber
\end{eqnarray}
\begin{prop} \label{prop:opt_two}
The optimal $\hw_l^t $ in (\ref{eq:obj_proximal}) is of the form $\hw_l^t = \alpha_l^t \p_l^t + \beta_l^t \q_l^t$, where 
$\alpha_l^t = \frac{\|\p_l^t \odot \d_l^{t-1} \odot \w_l^t\|_1}{\|\p_l^t \odot
\d_l^{t-1}\|_1}, 
\p_l^t = \I_{\alpha_l^t/2}^+(\w_l^t),
\beta_l^t = \frac{\|\q_l^t\odot \d_l^{t-1} \odot \w_l^t\|_1}{\|\q_l^t \odot \d_l^{t-1}\|_1}$,
and $\q_l^t = \I_{\beta_l^t/2}^-(\w_l^t)$.
\end{prop}

The exact and approximate solutions for 
$\alpha_l^t$ and $\beta_l^t$ 
can be obtained in a similar way as in Sections~\ref{sec:exact} and \ref{sec:approx}.
Details are in 
Appendix~\ref{apdx:two_scaling}.


\subsection{Extension to Low-Bit Quantization}
\label{sec:m-bit}

For $m$-bit 
quantization, we
simply change the set 
 $\mathcal{Q}$ 
of  desired quantized values in ~(\ref{eq:obj}) to
 one with $k=2^{m-1}-1$ quantized values. 
The optimization still contains a 
gradient descent step with adaptive learning rates like LAT, 
and a quantization step which can be solved efficiently by alternating minimization of
$(\alpha,
\b)$ (similar to the procedure in Algorithm~\ref{alg:approx})
using the following Proposition. 


\begin{prop} \label{prop:mbit_alt}
Let the optimal $\hat{\w}_l^t$ in (\ref{eq:ter_quantize}) be of the form $\alpha \b$.
	For a fixed $\b$,
	$\alpha = \frac{\|\b \odot \d_l^{t-1} \odot \w_l^t\|_1}{\|\b \odot  \d_l^{t-1}\|_1}$;
	whereas when
	$\alpha$ is
	fixed,
	$\b = \Pi_{\mathcal{Q}}(\frac{\w_l^t}{\alpha})$,
where $\mathcal{Q} =  \left\{-1, -\frac{k-1}{k}, \dots, -\frac{1}{k}, 0, \frac{1}{k}, \dots,
	\frac{k-1}{k}, 1\right\}$ for linear quantization and
	$\mathcal{Q}  = \left\{-1, -\frac{1}{2}, \dots, -\frac{1}{2^{k-1}}, 0, \frac{1}{2^{k-1}}, \dots,
	\frac{1}{2}, 1\right\}$ for logarithmic quantization.
\end{prop}


\section{Experiments}

In this section, we perform experiments 
on both
feedforward and recurrent neural networks.
The following methods are compared:
(i) the original full-precision network;
(ii) weight-binarized networks, including
		BinaryConnect 
		\citep{courbariaux2015binaryconnect},
		Binary-Weight-Network
		(BWN)~\citep{rastegari2016xnor},  and
		Loss-Aware Binarized network (LAB)~\citep{hou2017loss};
	(iii) weight-ternarized networks, including
		Ternary Weight Networks (TWN) \citep{li2016ternary},
		Trained Ternary Quantization (TTQ)\footnote{For TTQ, we follow the 
		\emph{CIFAR-10} 
		setting 
		in \citep{zhu2017trained}, and set $ \Delta_l^t = 0.005 \max(|\w_l^t|)
		$.}
		\citep{zhu2017trained},
		the proposed Loss-Aware Ternarized network with exact solution
	(LATe),
		approximate solution (LATa), and
		with two scaling parameters (LAT2e and LAT2a);
(iv) $m$-bit-quantized networks (where $m>2$), including
		DoReFa-Netm~\citep{zhou2016dorefa},
		the proposed loss-aware quantized network with linear quantization  (LAQm(linear)), and
logarithmic quantization (LAQm(log)).
Since weight quantization can be viewed as a form of regularization
\citep{courbariaux2015binaryconnect}, we do not use other regularizers such as dropout and weight decay.


\subsection{Feedforward Networks}
\label{sec:fnn}

In this section, we perform experiments with
the multilayer perceptron
(on the
\emph{MNIST}
data set)
and
convolutional neural networks
(on \emph{CIFAR-10}, \emph{CIFAR-100} and \emph{SVHN}).
For \emph{MNIST}, \emph{CIFAR-10}, and \emph{SVHN}, the setup is similar to that in
\citep{courbariaux2015binaryconnect,hou2017loss}. 
Details can be found in Appendix~\ref{apdx:expt_detail}.
For \emph{CIFAR-100},
we use $45,000$ images for training, another $5,000$ for validation, and the remaining $10,000$ for testing. 
The testing errors
are shown in
Table~\ref{tbl:fnn}.

\begin{table*}[ht]
	\centering
\caption{Testing errors (\%) on the feedforward networks. Algorithm with the lowest error in
each group is highlighted.}
	\label{tbl:fnn}
\begin{tabular}{ccc|c|c|c|c}
	\hline
	              &                        &                &  \emph{MNIST}  & \emph{CIFAR-10}  & \emph{CIFAR-100} &   \emph{SVHN}   \\ \hline
	 \multicolumn{2}{c}{no binarization}   & full-precision &      1.11      &      10.38       &      39.06       &      2.28       \\ \hline\hline
	              &                        & BinaryConnect  &      1.28      & \textbf{ 9.86 }  &      46.42       &      2.45       \\ \cline{3-7}
	   \multicolumn{2}{c}{binarization}    &      BWN       &      1.31      &      10.51       &      43.62       &      2.54       \\ \cline{3-7}
	              &                        &      LAB       &   {\bf 1.18}   &      10.50       &   {\bf 43.06}    &   {\bf 2.35}    \\ \hline\hline
	              &                        &      TWN       &      1.23      &      10.64       &      43.49       &      2.37       \\ \cline{3-7}	     
	              &       1 scaling        &     LATe     &      1.15      &      10.47       &      39.10       &   {\bf 2.30}    \\ \cline{3-7}
	ternarization &                        &     LATa     & \textbf{1.14 } & \textbf{ 10.38 } &      39.19       &   {\bf 2.30}    \\ \cline{2-7}
	              &                        &      TTQ       &      1.20      &      10.59       &      42.09       &      2.38       \\ \cline{3-7}
	              &       2 scaling        &    LAT2e     &      1.20      &      10.45       &      39.01       &      2.34       \\ \cline{3-7}
	              &                        &    LAT2a     &      1.19      &      10.48       &   {\bf 38.84}    &      2.35       \\ \hline\hline
	              &                        &   DoReFa-Net3   &      1.31      &      10.54       &      45.05       &      2.39       \\ \cline{3-7}
	\multicolumn{2}{c}{3-bit quantization} & LAQ3(linear)  &      1.20      &      10.67       &      38.70       &      2.34       \\ \cline{3-7}
	              &                        &   LAQ3(log)   &   {\bf 1.16}   &   {\bf 10.52}    & \textbf{38.50 }  & \textbf{ 2.29 } \\ \hline
\end{tabular}
\end{table*}

{\bf Ternarization:}
On \emph{MNIST}, \emph{CIFAR100} and \emph{SVHN}, the weight-ternarized 
networks perform better than weight-binarized networks,
and are comparable to the full-precision networks.
Among the weight-ternarized networks,
the proposed LAT and its variants
have the lowest errors.
On \emph{CIFAR-10}, LATa has similar performance as the full-precision network, but is
outperformed by BinaryConnect.

\begin{figure}[htbp]
\begin{center}
\subfigure[Training loss. \label{fig:train_cifar}]{\includegraphics[width=0.4\textwidth]{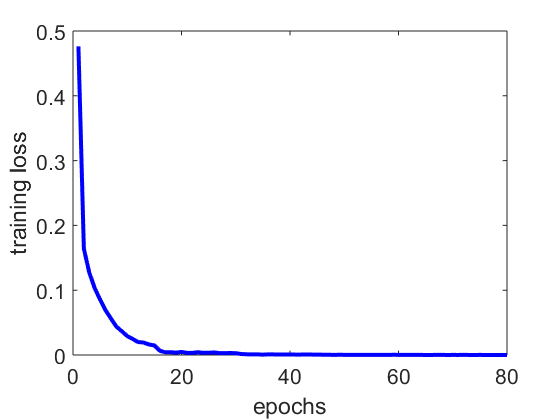}}
\subfigure[Scaling parameter $\alpha$.\label{fig:train_alpha}]{\includegraphics[width=0.4\textwidth]{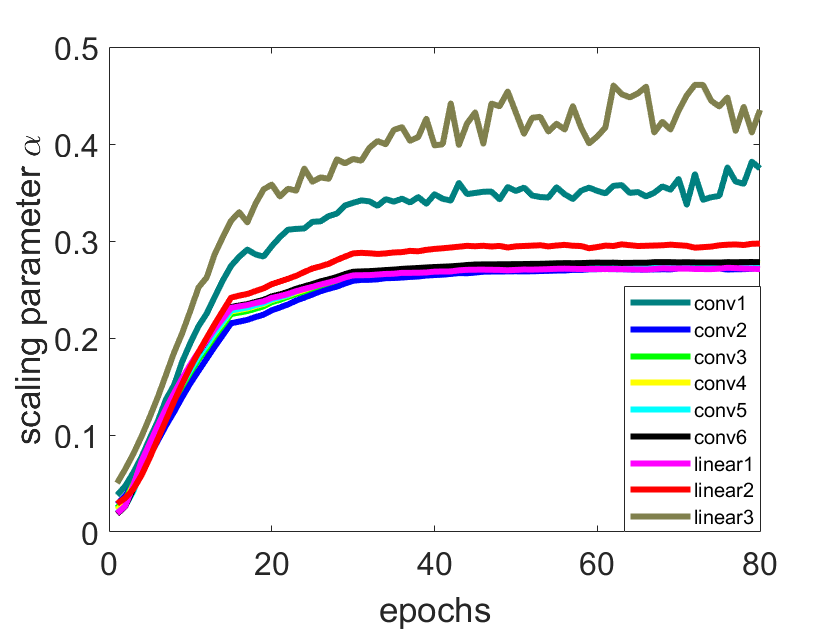}}
\vspace{-.1in} 
\caption{Convergence of the training loss
and scaling parameter by LATa
on \emph{CIFAR-10}.}
\label{fig:cifar}
\end{center}
\end{figure}

Figure~\ref{fig:train_cifar} shows convergence of the training loss for LATa on
\emph{CIFAR-10}, and Figure~\ref{fig:train_alpha} shows the scaling parameter obtained at
each CNN layer.  As can be seen, the scaling parameters for the first and last layers (conv1
and linear3, respectively) are larger than the others. This agrees with
the finding 
that, to maintain the activation variance and back-propagated gradients variance
during the forward and backward propagations,
the variance of the weights between the $l$th and $(l+1)$th layers
should roughly follow
$2/(n_l  + n_{l+1})$
\citep{glorot2010understanding}.
Hence, as the input 
and output layers
are small,
larger scaling parameters are needed for their high-variance weights.  

{\bf Using Two Scaling Parameters:}
Compared to TTQ,
the proposed LAT2 
always has better performance.
However, 
the extra flexibility 
of using two scaling parameters 
does not always translate to lower testing
error.
As can be seen,
it outperforms algorithms with one scaling parameter 
only on \emph{CIFAR-100}.
We speculate this is because  the 
capacities of deep networks are often larger than needed, and so the limited expressiveness of quantized weights
may not significantly deteriorate performance. Indeed, as pointed out in
\citep{courbariaux2015binaryconnect}, 
weight quantization is a form of regularization, and can contribute positively to the
performance.


{\bf Using More Bits:}
Among the 3-bit quantization algorithms,  the proposed scheme with 
logarithmic quantization has the best performance.   It also
outperforms the other quantization algorithms 
on \emph{CIFAR-100} and \emph{SVHN}.
However, as discussed above, more quantization flexibility is useful only when the
weight-quantized network does not have enough capacity.




\subsection{Recurrent Networks}
\label{sec:rnn}

In this section, we 
follow \citep{hou2017loss} and perform character-level language modeling
experiments on the
long short-term memory (LSTM) \citep{hochreiter1997long}.
The training objective is the cross-entropy loss 
over all target sequences.
Experiments are performed on three data sets:
(i) Leo Tolstoy's \emph{War and Peace}; (ii) source code of the \emph{Linux Kernel};
and
(iii) \emph{Penn Treebank} Corpus~\citep{taylor2003penn}.
For the first two, we follow the setting
in \citep{karpathy2015visualizing,hou2017loss}.
For \emph{Penn Treebank}, we follow the setting in \citep{mikolov2012context}.
In the experiment, we tried different initializations for TTQ and then report the best.
Cross-entropy values on the test set are shown in Table~\ref{tbl:rnn}.

\begin{table*}[htbp]	
\centering
\caption{Testing cross-entropy values on the LSTM. Algorithm with the lowest cross-entropy value in
	each group is highlighted.}
\label{tbl:rnn}
\begin{tabular}{ccc|c|c|c}
	\hline
	              &                         &                & \emph{War and Peace} & \emph{Linux Kernel}  & \emph{Penn Treebank} \\ \hline
	  \multicolumn{2}{c}{no binarization}   & full-precision &        1.268         &        1.326         &        1.083         \\ \hline\hline
	              &                         & BinaryConnect  &        2.942         &        3.532         &        1.737         \\ \cline{3-6}
	   \multicolumn{2}{c}{binarization}     &      BWN       &        1.313         &        1.307         &     {\bf 1.078}      \\ \cline{3-6}
	              &                         &      LAB       &     {\bf 1.291}      &   \textbf{1.305  }   &        1.081         \\ \hline\hline
	              &                         &      TWN       &        1.290         &        1.280         &        1.045         \\ \cline{3-6}
	              &        1 scaling        &      LATe      &        1.248         & \textbf{    1.256  } &        1.022         \\ \cline{3-6}
	ternarization &                         &      LATa      &        1.253         &        1.264         &        1.024         \\ \cline{2-6}
	              &                         &      TTQ       &        1.272         &        1.302         &        1.031         \\ \cline{3-6}
	              &        2 scaling        &     LAT2e      &        1.239         &        1.258         &        1.018         \\ \cline{3-6}
	              &                         &     LAT2a      &  \textbf{  1.245  }  &        1.258         &     {\bf 1.015}      \\ \hline\hline
	              &                         &  DoReFa-Net3   &        1.349         &        1.276         &        1.017         \\ \cline{3-6}
	\multicolumn{2}{c}{3-bit  quantization} &  LAQ3(linear)  &        1.282         &        1.327         &        1.017         \\ \cline{3-6}
	              &                         &   LAQ3(log)    &     {\bf 1.268}      &     {\bf 1.273}      &    \textbf{1.009}    \\ \hline\hline
	              &                         &  DoReFa-Net4   &        1.328         &        1.320         &        1.019         \\ \cline{3-6}
	\multicolumn{2}{c}{4-bit quantization}  & LAQ4 (linear)  &        1.294         &        1.337         &        1.046         \\ \cline{3-6}
	              &                         &   LAQ4 (log)   &     {\bf 1.272}      &     {\bf 1.319}      &     {\bf 1.016}      \\ \hline
\end{tabular}
\end{table*}

\begin{figure}[htbp]
	\begin{center}
		\subfigure[Training loss. \label{fig:train_warpeace}]{\includegraphics[width=0.40\textwidth]{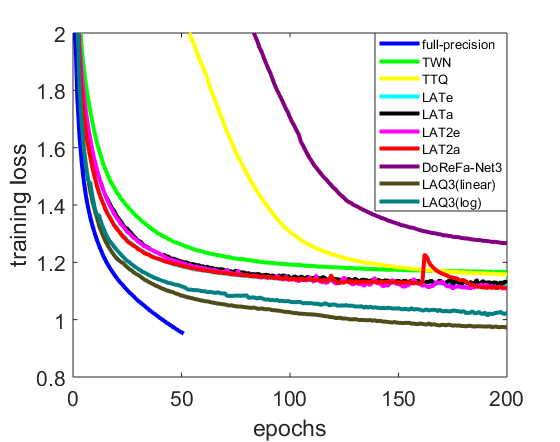}}
		\subfigure[Validation loss.\label{fig:val_warpeace}]{\includegraphics[width=0.40\textwidth]{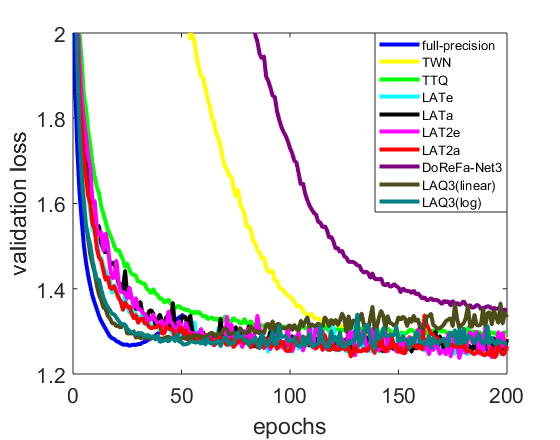}}
		\vspace{-.1in} 
\caption{Convergence of the training and validation losses on \emph{War and Peace}.}
		\label{fig:warpeace}
	\end{center}
\end{figure}

{\bf Ternarization:}
As in Section~\ref{sec:fnn},
the proposed LATe and LATa outperform the other weight ternarization schemes,
 and are even
better than the full-precision network on all  three data sets.
Figure~\ref{fig:warpeace} shows convergence of the training and validation losses on \emph{War and Peace}. 
Among the ternarization methods,  LAT and its variants
converge faster than both 
TWN
and 
TTQ.

{\bf Using Two Scaling Parameters:}
LAT2e and LAT2a
outperform TTQ
on all three data sets.
They also perform 
better than using one scaling parameter on 
\emph{War and Peace} and
\emph{Penn Treebank}.

\begin{figure}[htbp]
\begin{center}
\subfigure[Full-precision weights. \label{fig:full-3-bit}]{\includegraphics[width=0.245\textwidth]{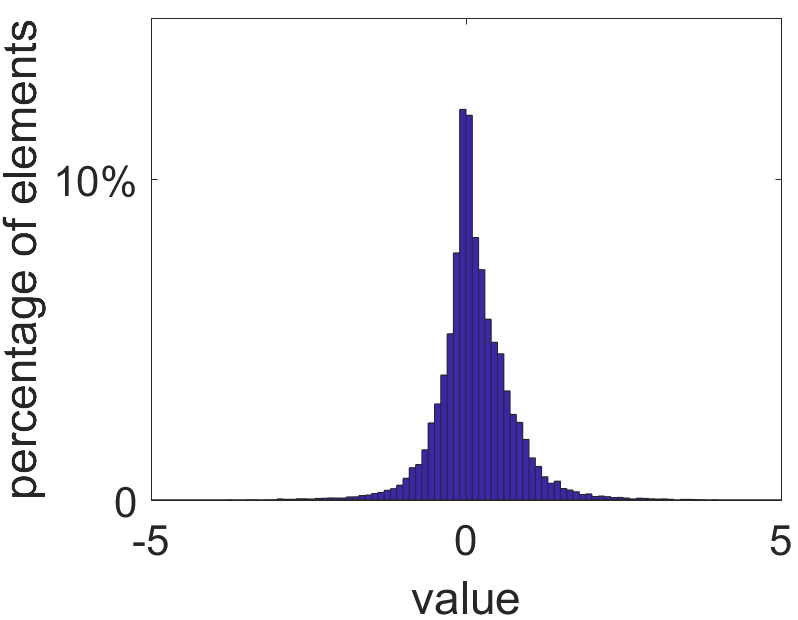}}
\subfigure[Quantized weights.
\label{fig:quantize-3-bit}]{\includegraphics[width=0.245\textwidth]{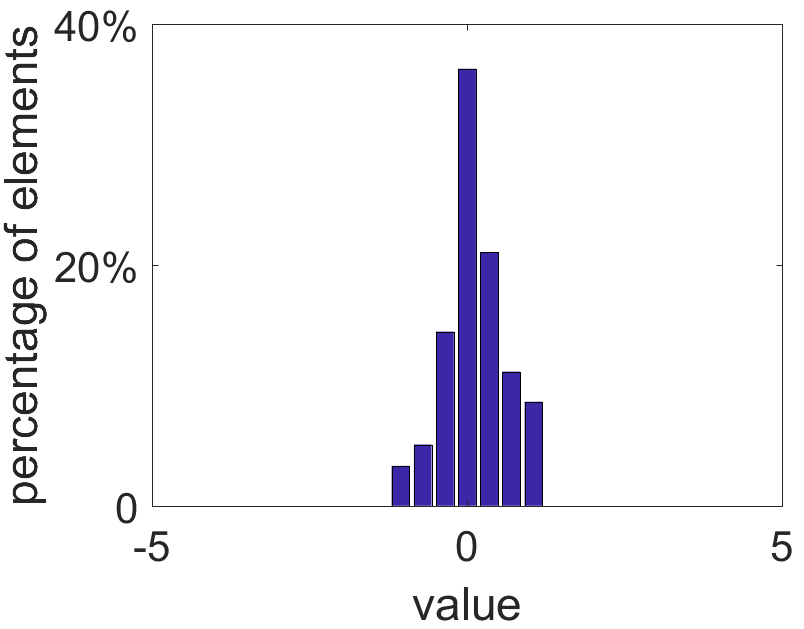}}
\subfigure[Full-precision weights. \label{fig:full-3-bit-log}]{\includegraphics[width=0.245\textwidth]{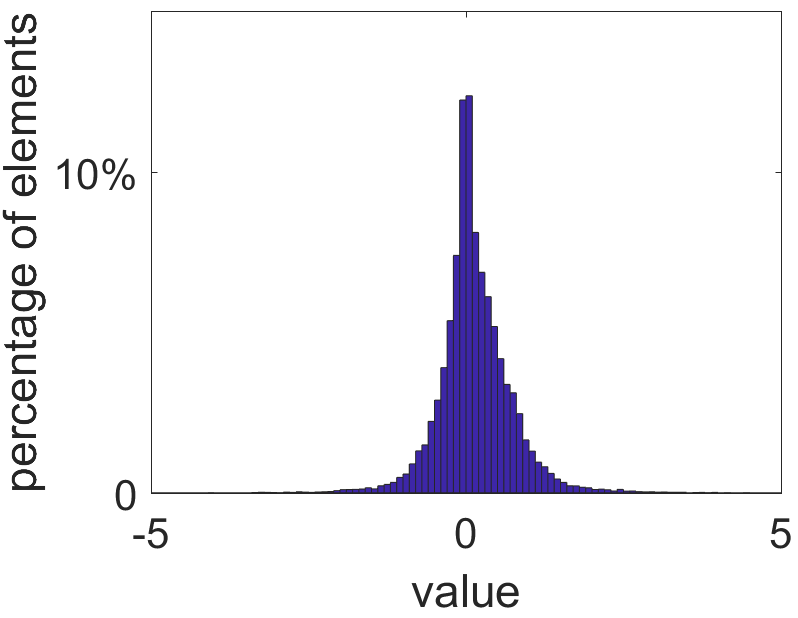}}
\subfigure[Quantized weights.\label{fig:quantize-3-bit-log}]{\includegraphics[width=0.245\textwidth]{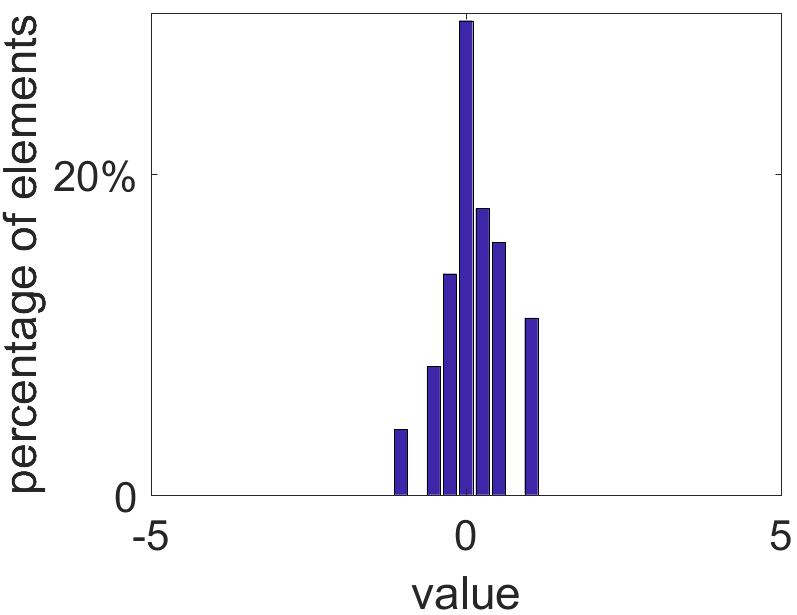}}
\vspace{-.1in} 
\caption{Distributions of the full-precision and 
LAQ3-quantized weights 
on \emph{War and Peace}.
Left ((a) and (b)): Linear quantization; Right ((c) and (d)): Logarithmic quantization.}
		\label{fig:weight}
	\end{center}
\end{figure}

{\bf Using More Bits:}
The proposed LAQ always outperforms
DoReFa-Net when 3 or 4 bits are used.
As noted in Section~\ref{sec:fnn}, using more 
bits 
does not necessarily yield better generalization performance,
and
ternarization (using 2 bits) yields the lowest validation loss on \emph{War and Peace} and \emph{Linux Kernel}.
Moreover, 
logarithmic quantization is better than linear quantization.
Figure~\ref{fig:weight} shows
distributions of  the
input-to-hidden 
(full-precision  and quantized)
weights
of the input gate trained 
after 20 epochs using LAQ3(linear) and LAQ3(log)
(results on the other weights are similar).
As can be seen, distributions 
of the full-precision weights
are bell-shaped.  Hence, logarithmic quantization can give finer
resolutions to many of the weights which have small magnitudes.


{\bf Quantized vs Full-precision Networks:}
The quantized networks often perform
better than the full-precision networks.
We speculate that this is because deep networks often have larger-than-needed capacities,
and so are less affected by the limited expressiveness of quantized weights. Moreover,
	low-bit quantization acts as regularization, and so contributes positively to the
	performance.


\section{Conclusion}

In this paper, we proposed a loss-aware weight quantization
algorithm that directly considers the 
effect
of 
quantization
on the loss.
The problem is solved using the proximal Newton algorithm.
Each iteration consists of  a preconditioned gradient descent step
and 
a quantization step that projects full-precision weights onto a set of quantized values.
For ternarization, 
an exact solution and an efficient approximate solution are provided.
The procedure is also extended to the use of different scaling parameters for the
positive and negative weights,
and to 
 $m$-bit (where $m>2$) quantization.
Experiments on both feedforward and recurrent networks show that the proposed quantization
scheme outperforms the current state-of-the-art.

\subsubsection*{Acknowledgments}

This research was supported in part by the Research Grants Council of the Hong Kong Special
Administrative Region
(Grant 614513).
We thank the
developers of Theano~\citep{2016arXiv160502688short},
Pylearn2~\citep{goodfellow2013pylearn2} and Lasagne.
We also thank NVIDIA for the gift of GPU card.

\bibliography{iclrRef}

\begin{thebibliography}{38}
\providecommand{\natexlab}[1]{#1}
\providecommand{\url}[1]{\texttt{#1}}
\expandafter\ifx\csname urlstyle\endcsname\relax
  \providecommand{\doi}[1]{doi: #1}\else
  \providecommand{\doi}{doi: \begingroup \urlstyle{rm}\Url}\fi

\bibitem[Boyd et~al.(2011)Boyd, Parikh, Chu, Peleato, and Eckstein]{boyd-11}
S.~Boyd, N.~Parikh, E.~Chu, B.~Peleato, and J.~Eckstein.
\newblock Distributed optimization and statistical learning via the alternating
  direction method of multipliers.
\newblock \emph{Foundations and Trends in Machine Learning}, 3\penalty0
  (1):\penalty0 1--122, 2011.

\bibitem[Courbariaux et~al.(2015)Courbariaux, Bengio, and
  David]{courbariaux2015binaryconnect}
M.~Courbariaux, Y.~Bengio, and J.~P. David.
\newblock {BinaryConnect}: Training deep neural networks with binary weights
  during propagations.
\newblock In \emph{Advances in Neural Information Processing Systems}, pp.\
  3105--3113, 2015.

\bibitem[Dauphin et~al.(2015)Dauphin, de~Vries, and
  Bengio]{dauphin2015equilibrated}
Y.~Dauphin, H.~de~Vries, and Y.~Bengio.
\newblock Equilibrated adaptive learning rates for non-convex optimization.
\newblock In \emph{Advances in Neural Information Processing Systems}, pp.\
  1504--1512, 2015.

\bibitem[Glorot \& Bengio(2010)Glorot and Bengio]{glorot2010understanding}
X.~Glorot and Y.~Bengio.
\newblock Understanding the difficulty of training deep feedforward neural
  networks.
\newblock In \emph{International Conference on Artificial Intelligence and
  Statistics}, pp.\  249--256, 2010.

\bibitem[Goodfellow et~al.(2013)Goodfellow, Warde-Farley, Lamblin, Dumoulin,
  Mirza, Pascanu, Bergstra, Bastien, and Bengio]{goodfellow2013pylearn2}
I.~J. Goodfellow, D.~Warde-Farley, P.~Lamblin, V.~Dumoulin, M.~Mirza,
  R.~Pascanu, J.~Bergstra, F.~Bastien, and Y.~Bengio.
\newblock Pylearn2: a machine learning research library.
\newblock Preprint, 2013.

\bibitem[Han et~al.(2015)Han, Pool, Tran, and Dally]{han2015learning}
S.~Han, J.~Pool, J.~Tran, and W.~J. Dally.
\newblock Learning both weights and connections for efficient neural network.
\newblock In \emph{Advances in Neural Information Processing Systems}, pp.\
  1135--1143, 2015.

\bibitem[Han et~al.(2016)Han, Mao, and Dally]{han2015deep}
S.~Han, H.~Mao, and W.~J. Dally.
\newblock Deep compression: Compressing deep neural network with pruning,
  trained quantization and {H}uffman coding.
\newblock In \emph{International Conference on Learning Representations}, 2016.

\bibitem[He et~al.(2016)He, Zhang, Ren, and Sun]{he2016deep}
K.~He, X.~Zhang, S.~Ren, and J.~Sun.
\newblock Deep residual learning for image recognition.
\newblock In \emph{International Conference on Computer Vision and Pattern
  Recognition}, pp.\  770--778, 2016.

\bibitem[Hochreiter \& Schmidhuber(1997)Hochreiter and
  Schmidhuber]{hochreiter1997long}
S.~Hochreiter and J.~Schmidhuber.
\newblock Long short-term memory.
\newblock \emph{Neural Computation}, pp.\  1735--1780, 1997.

\bibitem[Hou et~al.(2017)Hou, Yao, and Kwok]{hou2017loss}
L.~Hou, Q.~Yao, and J.~T. Kwok.
\newblock Loss-aware binarization of deep networks.
\newblock In \emph{International Conference on Learning Representations}, 2017.

\bibitem[Howard et~al.(2017)Howard, Zhu, Chen, Kalenichenko, Wang, Weyand,
  Andreetto, and Adam]{howard2017mobilenets}
A.~G. Howard, M.~Zhu, B.~Chen, D.~Kalenichenko, W.~Wang, T.~Weyand,
  M.~Andreetto, and H.~Adam.
\newblock {MobileNets}: Efficient convolutional neural networks for mobile
  vision applications.
\newblock Preprint arXiv:1704.04861, 2017.

\bibitem[Iandola et~al.(2016)Iandola, Han, Moskewicz, Ashraf, Dally, and
  Keutzer]{iandola2016squeezenet}
F.~N. Iandola, S.~Han, M.~W. Moskewicz, K.~Ashraf, W.~J. Dally, and K.~Keutzer.
\newblock Squeezenet: Alexnet-level accuracy with 50x fewer parameters and
  $<$0.5{MB} model size.
\newblock Preprint arXiv:1602.07360, 2016.

\bibitem[Karpathy et~al.(2016)Karpathy, Johnson, and
  Li]{karpathy2015visualizing}
A.~Karpathy, J.~Johnson, and F.~F. Li.
\newblock Visualizing and understanding recurrent networks.
\newblock In \emph{International Conference on Learning Representations}, 2016.

\bibitem[Kim et~al.(2016)Kim, Park, Yoo, Choi, Yang, and
  Shin]{kim2015compression}
Y.~D. Kim, E.~Park, S.~Yoo, T.~Choi, L.~Yang, and D.~Shin.
\newblock Compression of deep convolutional neural networks for fast and low
  power mobile applications.
\newblock In \emph{International Conference on Learning Representations}, 2016.

\bibitem[Kingma \& Ba(2015)Kingma and Ba]{kingma2014adam}
D.~Kingma and J.~Ba.
\newblock Adam: A method for stochastic optimization.
\newblock In \emph{International Conference on Learning Representations}, 2015.

\bibitem[Lebedev et~al.(2014)Lebedev, Ganin, Rakhuba, Oseledets, and
  Lempitsky]{lebedev2014speeding}
V.~Lebedev, Y.~Ganin, M.~Rakhuba, I.~Oseledets, and V.~Lempitsky.
\newblock Speeding-up convolutional neural networks using fine-tuned
  cp-decomposition.
\newblock Preprint arXiv:1412.6553, 2014.

\bibitem[LeCun et~al.(2015)LeCun, Bengio, and Hinton]{lecun2015deep}
Y.~LeCun, Y.~Bengio, and G.~Hinton.
\newblock Deep learning.
\newblock \emph{Nature}, 521\penalty0 (7553):\penalty0 436--444, 2015.

\bibitem[Lee et~al.(2014)Lee, Sun, and Saunders]{lee2014proximal}
J.~D. Lee, Y.~Sun, and M.~A. Saunders.
\newblock Proximal {N}ewton-type methods for minimizing composite functions.
\newblock \emph{SIAM Journal on Optimization}, 24\penalty0 (3):\penalty0
  1420--1443, 2014.

\bibitem[Leng et~al.(2017)Leng, Li, Zhu, and Jin]{leng2017extremely}
C.~Leng, H.~Li, S.~Zhu, and R.~Jin.
\newblock Extremely low bit neural network: Squeeze the last bit out with admm.
\newblock Preprint arXiv:1707.09870, 2017.

\bibitem[Li \& Liu(2016)Li and Liu]{li2016ternary}
F.~Li and B.~Liu.
\newblock Ternary weight networks.
\newblock Preprint arXiv:1605.04711, 2016.

\bibitem[Li et~al.(2017{\natexlab{a}})Li, De, Xu, Studer, Samet, and
  T.]{li2017training}
H.~Li, S.~De, Z.~Xu, C.~Studer, H.~Samet, and Goldstein T.
\newblock Training quantized nets: A deeper understanding.
\newblock In \emph{Advances in Neural Information Processing Systems},
  2017{\natexlab{a}}.

\bibitem[Li et~al.(2017{\natexlab{b}})Li, Kadav, Durdanovic, Samet, and
  Graf]{li2017pruning}
H.~Li, A.~Kadav, I.~Durdanovic, H.~Samet, and H.~P. Graf.
\newblock Pruning filters for efficient convnets.
\newblock In \emph{International Conference on Learning Representations},
  2017{\natexlab{b}}.

\bibitem[Lin et~al.(2016{\natexlab{a}})Lin, Talathi, and
  Annapureddy]{lin2016fixed}
D.~Lin, S.~Talathi, and S.~Annapureddy.
\newblock Fixed point quantization of deep convolutional networks.
\newblock In \emph{International Conference on Machine Learning}, pp.\
  2849--2858, 2016{\natexlab{a}}.

\bibitem[Lin et~al.(2016{\natexlab{b}})Lin, Courbariaux, Memisevic, and
  Bengio]{lin2015neural}
Z.~Lin, M.~Courbariaux, R.~Memisevic, and Y.~Bengio.
\newblock Neural networks with few multiplications.
\newblock In \emph{International Conference on Learning Representations},
  2016{\natexlab{b}}.

\bibitem[Mellempudi et~al.(2017)Mellempudi, Kundu, Mudigere, Das, Kaul, and
  Dubey]{mellempudi2017ternary}
N.~Mellempudi, A.~Kundu, D.~Mudigere, D.~Das, B.~Kaul, and P.~Dubey.
\newblock Ternary neural networks with fine-grained quantization.
\newblock Preprint arXiv:1705.01462, 2017.

\bibitem[Mikolov \& Zweig(2012)Mikolov and Zweig]{mikolov2012context}
T.~Mikolov and G.~Zweig.
\newblock Context dependent recurrent neural network language model.
\newblock \emph{IEEE Spoken Language Technology Workshop}, 12:\penalty0
  234--239, 2012.

\bibitem[Miyashita et~al.(2016)Miyashita, Lee, and
  Murmann]{miyashita2016convolutional}
D.~Miyashita, E.~H. Lee, and B.~Murmann.
\newblock Convolutional neural networks using logarithmic data representation.
\newblock Preprint arXiv:1603.01025, 2016.

\bibitem[Molchanov et~al.(2017)Molchanov, Tyree, Karras, Aila, and
  Kautz]{molchanov2017pruning}
P.~Molchanov, S.~Tyree, T.~Karras, T.~Aila, and J.~Kautz.
\newblock Pruning convolutional neural networks for resource efficient transfer
  learning.
\newblock In \emph{International Conference on Learning Representations}, 2017.

\bibitem[Novikov et~al.(2015)Novikov, Podoprikhin, Osokin, and
  Vetrov]{novikov2015tensorizing}
A.~Novikov, D.~Podoprikhin, A.~Osokin, and D.~P. Vetrov.
\newblock Tensorizing neural networks.
\newblock In \emph{Advances in Neural Information Processing Systems}, pp.\
  442--450, 2015.

\bibitem[Rakotomamonjy et~al.(2016)Rakotomamonjy, Flamary, and
  Gasso]{rakotomamonjy2016dc}
A.~Rakotomamonjy, R.~Flamary, and G.~Gasso.
\newblock {DC} proximal {N}ewton for nonconvex optimization problems.
\newblock \emph{IEEE Transactions on Neural Networks and Learning Systems},
  27\penalty0 (3):\penalty0 636--647, 2016.

\bibitem[Rastegari et~al.(2016)Rastegari, Ordonez, Redmon, and
  Farhadi]{rastegari2016xnor}
M.~Rastegari, V.~Ordonez, J.~Redmon, and A.~Farhadi.
\newblock {XNOR-Net}: {ImageNet} classification using binary convolutional
  neural networks.
\newblock In \emph{European Conference on Computer Vision}, 2016.

\bibitem[Szegedy et~al.(2015)Szegedy, Liu, Jia, Sermanet, Reed, Anguelov,
  Erhan, Vanhoucke, and Rabinovich]{szegedy2015going}
C.~Szegedy, W.~Liu, Y.~Jia, P.~Sermanet, S.~Reed, D.~Anguelov, D.~Erhan,
  V.~Vanhoucke, and A.~Rabinovich.
\newblock Going deeper with convolutions.
\newblock In \emph{International Conference on Computer Vision and Pattern
  Recognition}, pp.\  1--9, 2015.

\bibitem[Taylor et~al.(2003)Taylor, Marcus, and Santorini]{taylor2003penn}
A.~Taylor, M.~Marcus, and B.~Santorini.
\newblock The {P}enn treebank: An overview.
\newblock In \emph{Treebanks}, pp.\  5--22. Springer, 2003.

\bibitem[{Theano Development Team}(2016)]{2016arXiv160502688short}
{Theano Development Team}.
\newblock {Theano: A {Python} framework for fast computation of mathematical
  expressions}.
\newblock Preprint arXiv:1605.02688, 2016.

\bibitem[Vasilyev et~al.(2010)Vasilyev, Khoroshilova, and
  Antipin]{vasilyev2010extragradient}
F.~P. Vasilyev, E.~V. Khoroshilova, and A.~S. Antipin.
\newblock An extragradient method for finding the saddle point in an optimal
  control problem.
\newblock \emph{Moscow University Computational Mathematics and Cybernetics},
  34\penalty0 (3):\penalty0 113--118, 2010.

\bibitem[Zhang et~al.(2017)Zhang, Zhou, Lin, and Sun]{zhang2017shufflenet}
X.~Zhang, X.~Zhou, M.~Lin, and J.~Sun.
\newblock {ShuffleNet}: An extremely efficient convolutional neural network for
  mobile devices.
\newblock Preprint arXiv:1707.01083, 2017.

\bibitem[Zhou et~al.(2016)Zhou, Ni, Zhou, Wen, Wu, and Zou]{zhou2016dorefa}
S.~Zhou, Z.~Ni, X.~Zhou, H.~Wen, Y.~Wu, and Y.~Zou.
\newblock {DoReFa-Net}: Training low bitwidth convolutional neural networks
  with low bitwidth gradients.
\newblock Preprint arXiv:1606.06160, 2016.

\bibitem[Zhu et~al.(2017)Zhu, Han, Mao, and Dally]{zhu2017trained}
C.~Zhu, S.~Han, H.~Mao, and W.~J. Dally.
\newblock Trained ternary quantization.
\newblock In \emph{International Conference on Learning Representations}, 2017.

\end{thebibliography}
\bibliographystyle{iclr2018_conference}

\newpage
\appendix


\section{Proofs}


\subsection{Proof of Proposition~\ref{prop:two_step}}
\label{app:prop_opt}
	With $\w_l^t$ in (\ref{eq:ter_sgd}), the objective in (\ref{eq:obj_proximal}) can be
	rewritten as
	\begin{eqnarray*}
		\lefteqn{\nabla \ell(\hw^{t-1})^\top (\hw^t - \hw^{t-1}) + \frac{1}{2}(\hw^t - \hw^{t-1})^\top \D^{t-1} (\hw^t - \hw^{t-1})}
		\notag \\
		& = &  \frac{1}{2} \sum_{l=1}^{L}  \|\hw_l^t - (\hw_l^{t-1} -
		\nabla_l \ell(\hw^{t-1}) \oslash \d^{t-1}_l)\|_{\D_l^{t-1}}^2 + c_1
		\notag \\
		& = & \frac{1}{2} \sum_{l=1}^{L} \|\hw_l^t - \w_l^t\|_{{\D_l^{t-1}}}^2 + c_1
		\\
		& = & \frac{1}{2} \sum_{l=1}^{L} \sum_{i=1}^{n_l}  [\d_l^{t-1}]_i (\alpha_l^t [\b_l^t]_i
		- [\w_l^t]_i)^2 + c_1,
		\notag
	\end{eqnarray*}	
	where $c_1 = -\frac{1}{2} \|\nabla_l \ell(\hw^{t-1})
	\oslash \d^{t-1}_l\|_{\D_l^{t-1}} ^2 $ is independent of $\alpha_l^t$ and $\b_l^t$.


\subsection{Proof of Proposition~\ref{prop:ter_alt}}
\label{apdx:ter_alt}


To simplify notations, we drop the subscript and superscript. 
Considering one particular layer,  problem~(\ref{eq:ter_quantize}) is of the form:
\begin{eqnarray*}
& \min_{\alpha, \b} & 
\frac{1}{2} \sum_{i=1}^n  d_i (\alpha b_i - w_i)^2 \\
& \text{s.t.} & \alpha>0, b_i \in \{-1, 0,1\}.
\end{eqnarray*}
When $\alpha$ is fixed, 
\[
b_i = \arg \min_{b_i}\frac{1}{2}  d_i (\alpha b_i - w_i)^2 
= \frac{1}{2}   d_i \alpha^2(b_i - w_i/\alpha)^2
= \I_{\alpha/2}(w_i).
\]

When $\b$ is fixed,
\begin{eqnarray*}
	\alpha & = &\arg \min_{\alpha} \frac{1}{2} \sum_{i=1}^n  d_i (\alpha b_i - w_i)^2  \\
	& = &  \arg \min_{\alpha} \frac{1}{2} \|\b \odot \b \odot \d\|_1 \alpha^2 - \|\b \odot \d \odot \w\|_1 \alpha + c_2, \notag\\
	& = & \arg \min_{\alpha} \frac{1}{2}  \|\b \odot \b \odot \d\|_1 \left(\alpha -
	\frac{\|\b \odot \d \odot \w\|_1}{\|\b \odot \b \odot \d\|_1}
	\right)^2  -\frac{1}{2}  \frac{\|\b \odot \d \odot
		\w\|_1^2}{\|\b \odot \b \odot  \d\|_1} + c_2\\
	& = & \frac{\|\b \odot \d \odot \w\|_1}{\|\b \odot \b \odot \d\|_1}\\
	& = & \frac{\|\b \odot \d \odot \w\|_1}{\|\b \odot \d\|_1}.
\end{eqnarray*}


\subsection{Proof of Corollary~\ref{cor:twn} }
\label{apdx:twn}

When $\D_l^{t-1} = \lambda \I$,
i.e., the curvature is the same for all
dimensions in the $l$th layer, 
From Proposition~\ref{prop:ter_alt},
\[ \alpha_l^t = 
\frac{\|\b \odot \d_l^{t-1} \odot \w_l^t\|_1}{\|\b \odot  \d_l^{t-1}\|_1}
=\frac{\|\I_{\alpha_l^t/2}(\w^t_l) \odot \w_l^t\|_1}{\|\I_{\alpha_l^t/2}(\w^t_l) \|_1} = \frac{1}{\|\I_{\Delta_l^t}(\w^t_l)\|_1} \sum_{i: [\w^t_l]_i> \Delta_l^t }|[\w_l^t]_i|, \]
\[	\Delta_l^t
= \frac{1}{2}\frac{\|\I_{\alpha_l^t/2} \odot \w_l^t\|_1}{\|\I_{\alpha_l^t/2}\|_1}
= \arg \max_{\Delta>0} \frac{1}{\|\I_{\Delta}(\w^t_l)\|_1} \left(\sum_{i :[\w^t_l]_i >\Delta} |[\w_l^t]_i| \right)^2.  \]
This is the same as the TWN solution in~(\ref{eq:twn}).


\subsection{Proof of Proposition~\ref{prop:global_alpha}}
\label{app:global_alpha}

For simplicity  of notations, we drop the subscript and superscript. 
For each layer, we have an optimization problem of the form

\begin{eqnarray*}
\lefteqn{\arg \min_{\alpha}  \frac{1}{2} \sum_{i=1}^n  d_i (\alpha b_i - w_i)^2 }\\
	&= & \arg \min_{\alpha} \|\b \odot \b \odot \d\|_1 \left(\alpha -
	\frac{\|\b \odot \d \odot \w\|_1}{\|\b \odot \b \odot \d\|_1}
	\right)^2  -\frac{\|\b \odot \d \odot
		\w\|_1^2}{\|\b \odot \b \odot  \d\|_1}\\
	& = & \arg \min_{\alpha} \|\I_{\alpha/2}(\w) \odot \I_{\alpha/2}(\w) \odot \d\|_1 \left(\alpha -
	\frac{\|\I_{\alpha/2}(\w) \odot \d \odot \w\|_1}{\|\I_{\alpha/2}(\w) \odot \I_{\alpha/2}(\w) \odot \d\|_1}
	\right)^2  -\frac{\|\I_{\alpha/2}(\w) \odot \I_{\alpha/2}(\w) \odot
		\w\|_1^2}{\|\I_{\alpha/2}(\w) \odot \I_{\alpha/2}(\w) \odot  \d\|_1}\\
	& =& \arg \min_{\alpha} - \frac{\|\I_{\alpha/2}(\w) \odot \d \odot
	\w\|_1^2}{\|\I_{\alpha/2}(\w) \odot \d\|_1},
\end{eqnarray*}

where the second equality holds as $
\b = \I_{\alpha/2}(\w).
$
From (\ref{eq:tmp1}),
we have
\begin{eqnarray*}
\lefteqn{- \frac{\|\I_{\alpha/2}(\w) \odot \d \odot \w\|_1^2}{\|\I_{\alpha/2}(\w) \odot
\d\|_1}} \\
&= & - \frac{\|\I_{\alpha/2}(\w) \odot \d \odot \w\|_1}{\|\I_{\alpha/2}(\w) \odot \d\|_1} \cdot \frac{\|\I_{\alpha/2}(\w) \odot \d \odot \w\|_1}{\|\I_{\alpha/2}(\w) \odot \d\|_1} \cdot \|\I_{\alpha/2}(\w) \odot \d\|_1\\
& = & - 2c_j \cdot 2c_j \cdot [\text{cum}(\perm_{\s}(\d))]_j\\
& =& -2 c_j^2 \cdot [\text{cum}(\perm_{\s}(\d))]_j.
\end{eqnarray*}
Thus, the $\alpha$ 
that minimizes 
$ \frac{1}{2} \sum_{i=1}^n  d_i (\alpha b_i - w_i)^2 $
is 
$\alpha = 2 \arg \max_{c_j, j \in \mathcal{S}}  c_j^2 \cdot [\text{cum}(\perm_{\s}(\d))]_j$.

	
\subsection{Proof for Proposition~\ref{prop:opt_two}}
\label{apdx:opt_two}

For simplicity  of notations, we drop the subscript and superscript, and
consider the optimization
problem:
\begin{eqnarray*}\label{eq:ternarization_two}
& \min_{\alpha, \b} & \frac{1}{2} \sum_{i=1}^n  d_i ( \hat{w}_i - w_i)^2 \\
& \text{s.t.} & \hat{w}_i \in \{-\beta,0,+\alpha\}.
\end{eqnarray*}
Let $f(\hat{w}_i ) = (\hat{w}_i  - w_i)^2$. Then,
$f(\alpha) = (\alpha - w_i)^2, f(0) = w_i^2$, and $f(-\beta) = (\beta + w_i)^2$. 
It is easy to see that (i) if $w_i>\alpha/2, f(\alpha)$ is the smallest; (ii) if $w_i< -
\beta/2, f(-1)$ is the smallest;  (iii) if $-\beta/2 \leq w_i \leq \alpha/2, f(0)$ is the smallest.
In other words, the optimal $\hat{w}_i$ satisfies
\begin{equation*}
\hat{w}_i =  \alpha\I_{\alpha/2}^+(w_i) + \beta\I_{\beta/2}^-(w_i),
\end{equation*}
or equivalently,
$\hw  =  \alpha \p + \beta\q$,
where
$\p = \I_{\alpha/2}^+(\w)$, and $\q = \I_{\beta}^-(\w)$. 

Define $\w^+$ and $\w^-$ such that
$[\w^+]_i=
\begin{cases}
w_i  & w_i>0 \\
0 &\text{otherwise},
\end{cases}$
and 
$[\w^-]_i=
\begin{cases}
w_i & w_i<0 \\
0 & \text{otherwise}.
\end{cases}$.
Then,
\begin{equation}
\label{eq:tmp}
\frac{1}{2} \sum_{i=1}^n  d_i ( \hat{w}_i - w_i)^2 
= \frac{1}{2} \sum_{i=1}^n  d_i (\alpha p_i - w^+_i)^2 + \frac{1}{2} \sum_{i=1}^n  d_i (\beta q_i - w^-_i)^2.
\end{equation}
The objective 
in (\ref{eq:tmp}) has
two parts,
and each part can be viewed as a special case of the ternarization step in
Proposition~\ref{prop:two_step} (considering only with positive or negative weights).
Similar to the proof for Proposition~\ref{prop:ter_alt}, we can obtain that
the optimal $\hw= \alpha \p + \beta \q$ satisfies
\begin{eqnarray*}
	&\alpha = \frac{\|\p \odot \d \odot \w\|_1}{\|\p \odot \d\|_1}, 
	& \p = \I_{\alpha/2}^+(\w),
	\\
	& \beta = \frac{\|\q \odot \d \odot \w\|_1}{\|\q \odot \d\|_1},
	& \q = \I_{\beta/2}^-(\w).
\end{eqnarray*}

\subsection{Proof of Proposition~\ref{prop:mbit_alt}}
\label{apdx:mbit_alt}

For simplicity  of notations, we drop the subscript and superscript.
For each layer, we simply consider the optimization
problem:
\begin{eqnarray*}
& \min_{\alpha, \b} & \frac{1}{2} \sum_{i=1}^n  d_i (\alpha b_i - w_i)^2 \\
& \text{s.t.} & \alpha>0, b_i \in \mathcal{Q}.
\end{eqnarray*}
When $\alpha$ is fixed, 
\[
b_i = \arg \min_{b_i}\frac{1}{2}  d_i (\alpha b_i - w_i)^2 
= \frac{1}{2}   d_i \alpha^2(b_i - w_i/\alpha)^2
= \Pi_{\mathcal{Q}}\left(\frac{w_i}{\alpha}\right).
\]
When $\b$ is fixed,
\begin{eqnarray*}
\alpha & = &\arg \min_{\alpha} \frac{1}{2} \sum_{i=1}^n  d_i (\alpha b_i - w_i)^2  \\
& = &  \arg \min_{\alpha} \frac{1}{2} \|\b \odot \b \odot \d\|_1 \alpha^2 - \|\b \odot \d \odot \w\|_1 \alpha + c_2 \notag\\
& = & \arg \min_{\alpha} \frac{1}{2}  \|\b \odot \b \odot \d\|_1 \left(\alpha -
\frac{\|\b \odot \d \odot \w\|_1}{\|\b \odot \b \odot \d\|_1}
\right)^2  -\frac{1}{2}  \frac{\|\b \odot \d \odot
	\w\|_1^2}{\|\b \odot \b \odot  \d\|_1}\\
& = & \frac{\|\b \odot \d \odot \w\|_1}{\|\b \odot \b \odot \d\|_1}\\
& = & \frac{\|\b \odot \d \odot \w\|_1}{\|\b \odot \d\|_1}.
\end{eqnarray*}

\section{Loss-Aware Ternarization Algorithm (LAT) }
The whole procedure of LAT is shown in Algorithm~\ref{alg:whole}.
\label{apdx:whole}
\begin{algorithm}[h]
	\caption{Loss-Aware Ternarization (LAT)
		for training a feedforward neural network.}\label{alg:whole}
	\textbf{Input: }Minibatch 
	$\{(\x_0^t,\y^t)\}$,  current full-precision weights $\{\w^t_l\}$,
	first moment $\{\m^{t-1}_l\}$,
	second moment $\{\v^{t-1}_l\}$,
	and learning rate $\eta^t$. 
	\begin{algorithmic}[1]
		\STATE {\bf Forward Propagation}
		\FOR{$l=1$ to $L$}
		\STATE compute $\alpha^t_l$ and $\b_l^t$ using Algorithm~\ref{alg:exact} or \ref{alg:approx};
		\STATE rescale the layer-$l$ input: $\tilde{\x}^t_{l-1} = \alpha^t_l \x^t_{l-1}$;
		\STATE compute $\z^t_l$ with input  $\tilde{\x}^t_{l-1}$ and binary weight $\b^t_l$; 
		\STATE apply batch-normalization 
		and nonlinear activation to $\z^t_l$ to obtain $\x^t_l$; 
		\ENDFOR
		\STATE compute the loss $\ell$ using $\x^t_{L}$ and $\y^t$;
		\STATE {\bf Backward Propagation}
		\STATE initialize output layer's activation's gradient $\frac{\partial \ell}{\partial \x^t_{L}}$;
		\FOR{$l=L$ to $2$}
		\STATE compute $\frac{\partial \ell}{\partial \x^t_{l-1}}$ using $\frac{\partial
			\ell}{\partial \x^t_l}$, $\alpha^t_l$ and $\b^t_l$;
		\ENDFOR
		\STATE {\bf Update parameters using Adam}
		\FOR{$l=1$ to $L$}
		\STATE compute gradients $\nabla_l \ell(\hw^t)$ using $\frac{\partial \ell}{\partial \x^t_l}$ and $\x^t_{l-1}$;
		\STATE update first moment $\m^t_l = \beta_1 \m^{t-1}_l + (1-\beta_1)\nabla_l
		\ell(\hw^t)$;
		\STATE update second moment $\v^t_l = \beta_2 \v^{t-1}_l + (1-\beta_2)(\nabla_l \ell(\hw^t) \odot \nabla_l \ell(\hw^t))$;
		\STATE compute unbiased first moment $\hat{\m}^t_l = \m^t_l/(1-\beta_1^t)$;
		\STATE compute unbiased second moment $\hat{\v}^t_l = \v^t_l/(1-\beta_2^t)$;
		\STATE compute current curvature matrix $\d^t_l = \frac{1}{\eta^t}\left(\epsilon \bm{1}+\sqrt{\hat{\v}^t_l}
		\right)$;
		\STATE update full-precision weights $\w^{t+1}_l = \w^t_l -  \hat{\m}^t_l \oslash \d^t_l$;
		\STATE update learning rate $\eta^{t+1} = \text{UpdateLearningrate} (\eta^t, t+1)$;
		\ENDFOR
	\end{algorithmic}
\end{algorithm}


\section{Exact and Approximate Solutions for Ternarization with Two Scaling Parameters}
\label{apdx:two_scaling}

Let there be $n_1$ positive elements and $n_2$ negative elements in $\w_l$.
For a $n$-dimensional vector $\x =[x_1, x_2, \dots, x_n]$,  define
$\text{inverse}(\x) = [x_n, x_{n-1}, \dots , x_1]$.
As is shown in (\ref{eq:tmp}), the objective can be separated into two parts,
and each part can be viewed as a special case of ternarization step in Proposition~\ref{prop:two_step}, dealing only with positive or negative weights.
Thus the exact and approximate solutions for $\alpha_l^t$ and $\beta_l^t$ can separately be derived in a similar way as that of using one scaling parameter.
The exact and approximate solutions for 
$\alpha_l^t$ and $\beta_l^t$ for layer-$l$ at the $t$th time step are shown in Algorithms~\ref{alg:2} and \ref{alg:2_approx}.

\begin{algorithm}[htbp]
	\caption{Exact solver for $\hw_l^t$ with two scaling parameters.}\label{alg:2}
	\begin{algorithmic}[1]
		\STATE{\bf Input:} full-precision weight $\w_l^t$, diagonal entries of the approximate Hessian $\d_l^{t-1}$.
		\STATE $\s_1 = \arg \text{sort} (\w_l^t)$;
		\STATE $\c_1 = \text{cum}(\text{perm}_{\s_1}(|\d_l^{t-1} \odot \w_l^t |)) \oslash \text{cum}(\text{perm}_{\s_1}(|\d_l^{t-1} |)) \oslash 2$;
		\STATE $\S_1 = \text{find}[([\text{perm}_{\s_1}(\w_l^t )]_{[1:(n_1-1)]}-[\c_1]_{[1:(n_1-1)]}) \odot [\text{perm}_{\s_1}(\w_l^t)]_{[2:n_1]} - [\c_1]_{[1:n_1-1]}) < 0 )$;
		\STATE $\alpha_l^t = 2\arg \max_{c_i, i \in \S_1} [\c_1]_i^2 \cdot [\text{cum}(\text{perm}_{\s_1}(|\d_l^{t-1}|))]_i$;
		\STATE $\p_l^t = \I_{\alpha/2}^+(\w_l^t)$;
		\STATE $\s_2 = \text{inverse}(\s_1)$;
		\STATE $\c_2 = \text{cum}(\text{perm}_{\s_2}(|\d_l^{t-1} \odot \w_l^t |)) \oslash \text{cum}(\text{perm}_{\s_2}(|\d_l^{t-1} |)) \oslash 2$;
		\STATE $\S_2 = \text{find}(([-\text{perm}_{\s_2}(\w_l^t)]_{[1:(n_2-1)]}-[\c_2]_{[1:(n_2-1)]}) \odot ([-\text{perm}_{\s_2}(\w_l^t)]_{[2:n_2]} - [\c_2]_{[1:n_2-1]}) < 0 )$;
		\STATE $\beta_l^t = 2\arg \max_{c_i, i \in \S_2} [\c_2]_i^2 \odot [\text{cum}(\text{perm}_{\s_2}(|\d_l^{t-1}|))]_i$;
		\STATE $\q_l^t = \I_{\beta/2}^-(\w_l^t)$;
		\STATE{\bf Output:}  $\hw_l^t = \alpha_l^t \p_l^t + \beta_l^t \q_l^t$.
	\end{algorithmic}
\end{algorithm}

\begin{algorithm}[htbp]
	\caption{Approximate solver for $\hw_l^t$ with two scaling parameters}\label{alg:2_approx}
	\begin{algorithmic}[1]
		\STATE{\bf Input:} $\b_l^{t-1}$, full-precision weight $\w_l^t$,  and diagonal entries of approximate Hessian $\d_l^{t-1}$.
		\STATE{\bf Initialize:} $\alpha = 1.0, \alpha_{\text{old}} = 0.0, \beta = 1.0, \beta_o = 0.0, \b=\b_l^{t-1}, \p= \I_0^+(\b), \q=\I_0^-(\b), \epsilon=10^{-6} $.	
\WHILE{$|\alpha - \alpha_{\text{old}}|>\epsilon$ \textbf{and} $|\beta -
\beta_{\text{old}}|>\epsilon$}
		\STATE $\alpha_{\text{old}} = \alpha$, $\beta_{\text{old}} = \beta$;
		\STATE $\alpha = \frac{\|\p \odot \d_l^{t-1} \odot \w_l^t\|_1}{\|\p \odot \d_l^{t-1}\|_1}$;
		\STATE $\p = \I_{\alpha/2}^+(\w_l^t)$;
		\STATE $\beta = \frac{\|\q \odot \d_l^{t-1} \odot \w_l^t\|_1}{\|\q  \odot \d_l^{t-1}\|_1}$;
		\STATE $\q = \I_{\beta/2}^-(\w_l^t)$;
		\ENDWHILE
		\STATE{\bf Output:} $\hw_l^t = \alpha \p + \beta \q$.
	\end{algorithmic}
\end{algorithm}

\section{Experimental Details}
\label{apdx:expt_detail}

\subsection{Setup for Feedforward Networks}

The setup for the four data sets are as follows:
\begin{enumerate}
\item  \emph{MNIST}:
This contains $28 \times 28$ gray images from 10 digit classes.  We use $50,000$ images for
training, another $10,000$ for validation, and the remaining $10,000$  for testing.  
		We use the 4-layer model:
		\[ 784FC-2048FC-2048FC-2048FC-10SVM, \]
		where $FC$ is a fully-connected layer, and $SVM$ is a $\ell_2$-SVM output layer using the square hinge
		loss.
		Batch normalization with a minibatch size $100$,
		is used to accelerate learning. 
		The maximum number of epochs is $50$.
		The learning rate 
		starts at $0.01$,
		and decays by a factor of $0.1$ at epochs $15$ and $25$.
		
\item \emph{CIFAR-10}: This contains $32 \times 32$ color images from 10 object classes.  We
use $45,000$ images for training, another $5,000$ for validation, and the remaining $10,000$ 
			for testing. 
		The images are preprocessed with global contrast normalization and ZCA whitening.
		We 
		use the VGG-like architecture:
		\[
		(2\times 128C3)-MP2-(2 \times 256C3)-MP2-(2 \times 512C3)-MP2-(2 \times 1024FC)-10SVM,
		\]
		where $C3$ is a $3 \times 3$ ReLU convolution layer, and $MP2$ is a $2 \times 2$ max-pooling layer. 
		Batch normalization with a minibatch size of $50$, is used.
		The maximum number of epochs is $200$.
		The learning rate for 
		the weight-binarized network starts at $0.03$ while for 
		all the other networks starts at $0.002$, and decays by a factor of $0.5$ after every 15 epochs. 
		
\item \emph{CIFAR-100}: This contains $32 \times 32$ color images from 100 object classes.
We use $45,000$ images for training, another $5,000$ for validation, and the remaining $10,000$ for testing. 
		The images are preprocessed with global contrast normalization and ZCA whitening.
		We 
		use the VGG-like architecture:
		\[
		(2\times 128C3)-MP2-(2 \times 256C3)-MP2-(2 \times 512C3)-MP2-(2 \times 1024FC)-100SVM.
		\]
		Batch normalization with a minibatch size of $100$, is used.
		The maximum number of epochs is $200$.
		The learning rate 
		starts at $0.0005$, and decays by a factor of $0.5$ after every 15 epochs. 	
		
\item \emph{SVHN}: This contains $32\times 32$ color images from 10 digit classes.  We use
$598,388$ images for training, another $6,000$ for validation, and the remaining $26,032$ 
			for testing.
		The images are preprocessed with global and local contrast normalization.
		The model used is:
		\[
		(2\times 64C3)-MP2-(2 \times 128C3)-MP2-(2 \times 256C3)-MP2-(2 \times 1024FC)-10SVM. 
		\]
		Batch normalization with a minibatch size of $50$, is used.
		The maximum number of epochs is $50$. 
The learning rate 
starts at $0.001$ 
		for the weight-binarized network, and
$0.0005$
for the other networks.
		It then decays by a factor of $0.1$ at epochs $15$ and $25$.
	\end{enumerate}

\subsection{Setup for Recurrent Networks}

The setup for the three data sets are as follows:
\begin{enumerate}
\item Leo Tolstoy's \emph{War and Peace}: It consists of 3258K characters of almost
entirely English text with minimal markup and a vocabulary size of $87$. We use the same training/validation/test set split as in \citep{karpathy2015visualizing,hou2017loss}.

\item The source code of the \emph{Linux Kernel}: This consists of 621K characters and 
a vocabulary size of $101$. We use the same training/validation/test set split as in \citep{karpathy2015visualizing,hou2017loss}.

\item The 
{\em Penn Treebank} 
data set 
\citep{taylor2003penn}:
This has been frequently used 
for language modeling.
It contains 50 different characters, including English characters, numbers, and
punctuations. We follow the setting in~\citep{mikolov2012context}, with 5,017K characters for training, 393K 
for validation, and 442K characters for testing. 
\end{enumerate}

We use a one-layer LSTM with $512$ cells.
The maximum number of epochs is $200$, 
and the number of time steps
is $100$.
The initial learning rate 
is 
$0.002$.
After $10$ epochs,
it is decayed 
by a factor of $0.98$ after each epoch.
The weights are initialized uniformly in $[−0.08, 0.08]$.
After each iteration, the gradients are clipped to the range $[-5,5]$.
All the updated weights are clipped to $[-1, 1]$ for binarization and ternarization methods, but not for $m$-bit (where $m>2$) quantization methods.
\end{document}